\documentclass[conference]{IEEEtran}
\IEEEoverridecommandlockouts
\usepackage{tabularx}
\usepackage{ragged2e}
\usepackage{cite}
\usepackage{amsmath,amssymb,amsfonts}
\usepackage{algorithmic}
\usepackage{graphicx}
\usepackage{textcomp}
\usepackage{float}
\usepackage{xcolor}
\def\BibTeX{{\rm B\kern-.05em{\sc i\kern-.025em b}\kern-.08em
    T\kern-.1667em\lower.7ex\hbox{E}\kern-.125emX}}
\begin{document}

\title{Contextual Code Switching for Machine Translation using Language Models\\

}

\author{
    \IEEEauthorblockN{
        Arshad Kaji
    }
            \IEEEauthorblockA{
                Mumbai, India \\
                arshadkaji0412@gmail.com
            }
\and
    \IEEEauthorblockN{
        Shah Manan Vinod
    }
        \IEEEauthorblockA{
            Mumbai, India \\
            shah.manan.vinod@gmail.com}
}

\maketitle

\begin{abstract}
Large language models (LLMs) have exerted a considerable impact on diverse language-related tasks in recent years. Their demonstrated state-of-the-art performance is achieved through methodologies such as zero-shot or few-shot prompting. These models undergo training on extensive datasets that encompass segments of the Internet and subsequently undergo fine-tuning tailored to specific tasks. Notably, they exhibit proficiency in tasks such as translation, summarization, question answering, and creative writing, even in the absence of explicit training for those particular tasks.  While they have shown substantial improvement in the multilingual tasks their performance in the code switching, especially for machine translation remains relatively uncharted. In this paper, we present an extensive study on the code switching task specifically for the machine translation task comparing multiple LLMs. Our results indicate that despite the LLMs having promising results in the certain tasks, the models with relatively lesser complexity outperform the multilingual large language models in the machine translation task. We posit that the efficacy of multilingual large language models in contextual code switching is constrained by their training methodologies. In contrast, relatively smaller models, when trained and fine-tuned on bespoke datasets, may yield superior results in comparison to the majority of multilingual models.
\end{abstract}


\begin{figure*}
    \centering
    \includegraphics[width=0.7\textwidth]{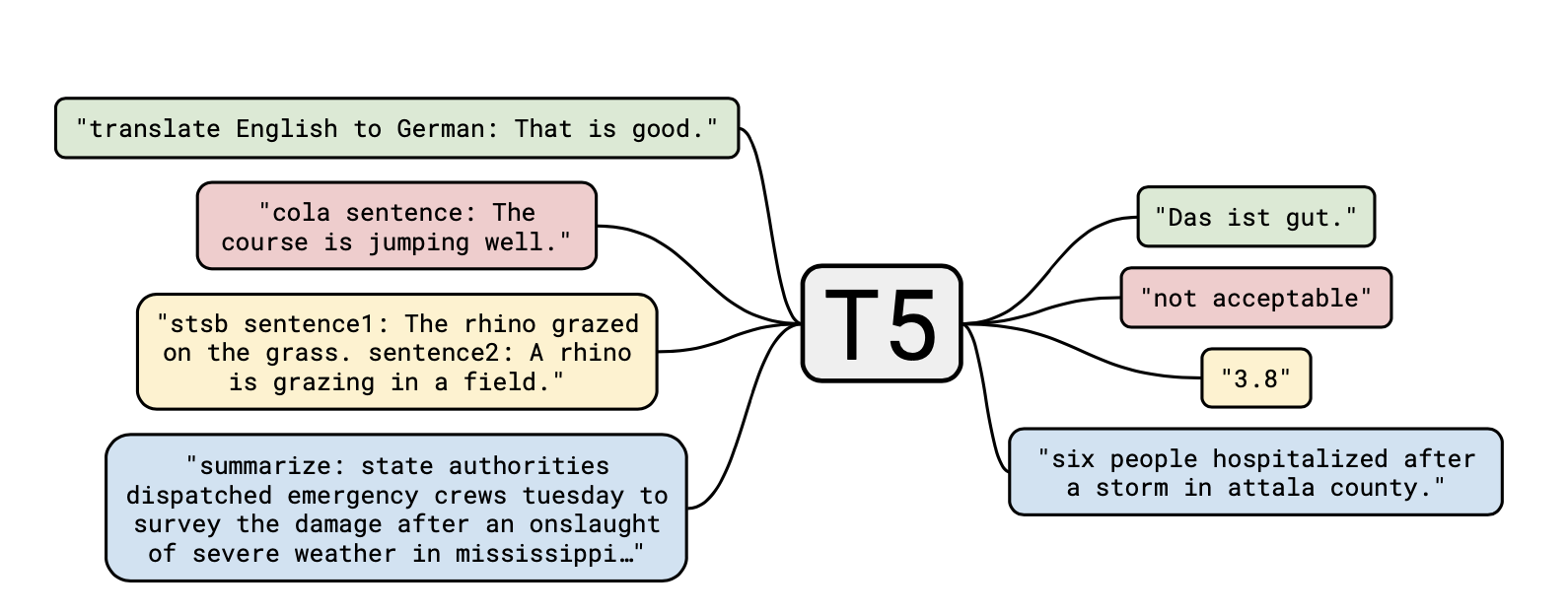}
    \caption{T5 Model Illustration}
    \small
    The depiction of the T5 model in \cite{b17} underscores its unified architecture applicable to diverse language tasks. Our investigation reveals that the T5-based model, owing to its unified structure, readily lends itself to effective fine-tuning on custom datasets. Notably, optimal outcomes were observed with T5-small, surpassing performance from larger language models.
\end{figure*}

\section{Introduction}

In multicultural societies like India, individuals frequently do not restrict themselves to a single language in order to overcome language barriers. Instead, they seamlessly switch between two or more languages during conversations, a phenomenon commonly referred to as code-switching (CSW)\cite{b1}. In general, due to the vast variety of languages people often use one or more languages to bridge the language barrier. This is more evident on social media sites. Hinglish, a mixture of Hindi and English is a predominant mode of conversation on online sites \cite{b2}. 

Recent advancements in Large Language Models (LLMs) have exhibited a substantial impact on diverse language tasks, particularly within the realm of multilingual settings, leveraging few-shot learning methods \cite{b3}. It is noteworthy to acknowledge that the task of code switching is typically not addressed by these language models, primarily due to the inherent challenges associated with acquiring code-switched data.However this field is getting increasingly explored seen by \cite{b4}.

The main challenge in code switching task is scarcity of the data. LLM optimisation for code switching is harder on the scarce data available today as well as colloquial characteristic of code-switching \cite{b5}. The study revealed that, LLMs such as InstructGPT \cite{b6} and ChatGPT outperform public models like BLOOMZ \cite{b7} and Flan-T5-XXL \cite{b8} in generating natural-sounding code switching texts revealed by \cite{b9}. The main limitations for these models to perform any code switching task is due to the fact that these models are trained on the multiple individual language data sets. To address the challenge of code-switching in machine translation, two distinct methodologies have been identified in prior research, as documented by \cite{b10}. The first method involves the utilization of a Language Identification (LID) translation model, employing a translation approach. The second approach entails the use of a separate translation model for word-based translation, integrated with a final model designed to enhance contextual understanding. 

Due to high level of complexity and scarcity of the code-switching text data, it is harder to fine tune LLMs for machine translation tasks. Our goal with this project is to create a best suited model which can convert Hinglish (Hindi + English) code switched text into English. We train a substantially smaller language model to mitigate the complexity of large language models and fine tune it to give results for the Hinglish to English translation. Essentially, our model will be provided with sentences such as "Hi, aapko movie pasand ayi?" with the objective of translating it into "Hi, did you like the movie?". Our emphasis is specifically on facilitating Hindi+English to English conversion, excluding consideration for the reverse transformation.

\section{Approach}

Our overarching objective is to construct a model that is not only straightforward to train but also possesses an enhanced grasp of contextual information within the language. While numerous Large Language Models (LLMs) already exist, showcasing state-of-the-art contextual understanding for language-related tasks, fine-tuning them on a limited amount of Hinglish data poses considerable challenges. 

We chose both multi-lingual as well as singlish (only English) models to experiment with. While choosing any model, considered their size, the architecture and their training techniques. We majorly focused on available fine-turning techniques and their compatibility \cite{b11} for such models. Models with masked language modeling (MLM) \cite{b11} pre-training methods are used for our tasks. We include models with both encoder and decoder end of the transformer architecture to get full text to text functionality. 

Code switching for machine translation is dissimilar to the standard machine translation tasks. Working with translation models would have inferior performance if used without fine-tuning as shown in \cite{b19}. Hence, instead of confining ourselves with only translation as well as multi-lingual models we tried different types of models with different training methods. 

The following concise points delineate the criteria that informed our model selection process:-

\begin{itemize}
    \item Models which are small enough to train and fine tune on Hinglish data.
    \item Models specifically designed for text-to-text language tasks, featuring both encoder and decoder components.
    \item Preference for smaller-sized models, emphasizing ease of fine-tuning over architectural intricacies and training methodologies.
    \item Acknowledgment of the dissimilarity between translation models trained on individual language data and the unique challenges posed by code-switching data. 
    \item Recognition that while multi-lingual Large Language Models (LLMs) exhibit some capacity for code-switching, their complex fine-tuning process may result in suboptimal performance compared to smaller, finely tuned models.
\end{itemize}

\section{Experimentation}
Table 1 represents the models we experimented with and their sizes. These contain varied range of models trained on different types of tasks. We took the smaller models that we could train and worked with Hinglish data for them. Some models which are larger in size, and which are not feasible to train on our limited resources, we used LoRA method \cite{b13} to fine-tune. 

LoRA method allows us to fine tune even the models with larger sizes on custom data by reducing the number of trainable parameters. For the models which are size above 100 million parameters, we used LoRA to train only 2-3\% of the actual model size. Thus, saving considerable amount of the compute power. For models BART \cite{b14}, Flan T5 base \cite{b8}, we used LoRA method for fine tuning. The amount of trainable parameter space can be defined by the rank of the LoRA matrix. We experimented with multiple ranks ranging from 8 to 32 on each of our models. Given their larger sizes and complexity, we found their performance to be lower compared to the smaller fully trained models.

Models such as Flan T5 small \cite{b8} which are smaller in size gave us promising results compared to other models. Original T5 model is a unified text to text model which is designed to perform multiple types of tasks such as translation, summarisastion, question-answering etc. shown in figure 1. By selecting the smaller model size which has only 77 million parameters we were able to train Flan T5 small on our hinglish data. And achieved best results compared to other models.

We also compared our models against the ChatGPT which is a RLHF fine tuned model \cite{b6}. Based on our results, after manual evaluation we found the consistency of ChatGPT to be lower compared to fully trained models. 

\begin{table}
  \centering
  \caption{Model Size Comparison}
  \begin{tabular}{|c|c|}
    \hline
    \textbf{Model} & \textbf{Parameter Size} \\
    \hline
    Flan T5 small & 77M \\
    BART base & 139M \\
    Flan T5 base & 248M \\
    \hline
  \end{tabular}
  \label{tab:model_comparison}
\end{table}


\begin{table*}
\centering    
    \caption{Example Translations}
    \begin{tabularx}{\textwidth}{|X|X|X|X|}
        \hline
            \textbf{Input (Code Switched - Text)} & \textbf{Expected Output (Ground Truth)} & \textbf{Zero-Shot Model Results(Flan T5 Small)} & \textbf{Proposed Model Results} \\ \hline
            yeh tho biographical movie ke tarah hein, mark zukerberk ke baare mein hein kya jo facebook create kiya  & It appears to be like a biographical movie about Mark Zuckerberg and when he creates Facebook. & yeh tho biographical movie to tarah hein, mark zu & It is kind of a biographical movie about Mark Zuckerberg who was in it's \\ \hline
            haan yar. mein bhi is movie ke baare mein suna hoon, kahi saal ke pehle aaya tha na. & Ohh yes I remember hearing about that movie, but it was a long time ago it feels like. & i'm not a movie maker, but i'm a movie maker & Yes. I've heard of it, it was a while back when it arrived. \\ \hline
            Rotten Tomatoes ne toh 96\% deya tha aur yeh movie tho 2010 ka hein & Rotten Tomatoes has rated it 96\% so that's decent, and it is a movie from 2010 & Rotten Tomatoes has rated this movie 96\%, and it's & Rotten Tomatoes gave it a 96\% and it was a movie from \\ \hline
            haan worth watching hein & may be worth watching! & hein worth watching & Yes, it's worth watching \\ \hline
            haan muje bhi aise hi lagtha, lekin personally muje facebook itna bhi pasand nahi hein. & Yes I think it could be, although I personally do not like facebook as a company. & I am not a fan of the sexy sexy girl, but & Yeah I think so too, but I personally don't really like facebook so much. \\ \hline
    \end{tabularx}
\end{table*}
\section{Dataset}
Hinglish dataset is scars compared to other code switch data. For our task of machine translation, we used \cite{b14} which has 8,060 pairs in the training split, 942 pairs in validation and 960 pairs in the test split. For the fine tuning of the larger models we used \cite{b16} which contains more than 8 thousand samples from hinglish to english. Both of these dataset contain both human as well as machine generated samples. 

To enhance the model, we introduced some noise in these datasets. Hinglish is not a standard language compared to Hindi or English. Hence people use their convenient version of the word. We found words such as "Toh", "Kyun", "Han" from Hindi can be used as "To", "Kyu", "Ha"; removing or adding phonetics in the end based on the formality of the context. Variation in spellings may lead to complete difference in the context in Hinglish. For some samples, word "To" which represents cause-and-effect in Hindi language context can be miss interpreted as "To" in English inside the code switched Hinglish data.  We added only the bilingual data (Hinglish) to our dataset. Words other than English or Hindi were removed from the dataset before training the model.

\section{Results}
\begin{table}[H]
  \centering
  \caption{BLEU Score Evaluation}
  \begin{tabular}{|c|c|}
    \hline
    \textbf{Datasets} & \textbf{BLEU Score} \\
    
    \hline
    Hinglish TOP Dataset\cite{b20} & 65.81 \\
    HinGE\cite{b15} & 29.2\\
    findnitai/english-to-hinglish\cite{b18} & 19.03 \\
    cmu hinglish dog\cite{b16} & 13.46\\
    \hline
  \end{tabular}
  \label{tab:model_comparison}
\end{table}

We have made use of BLEU score \cite{b19} for text-to-text tasks, as we are performing machine translations, because it provides a quantitative measure of the similarity between a generated text and reference texts. It also helps evaluate the quality of machine-generated translations by assessing how well the generated text matches human references. Its wide acceptance as a benchmark offers a practical and automated evaluation of the quality of generated code-switched text, notably in tasks like machine translation. However, BLEU has limitations, as it primarily relies on n-gram precision and doesn't capture semantic nuances or fluency. It may favor literal translations and may not align with human judgments on translation quality. Using BLEU in isolation can lead to over-optimization for certain linguistic patterns without genuine comprehension of meaning.

\section{Conclusion}
We studied multiple types of the language models for machine translation on code switched data. Models with relatively lesser complexity outperform the multilingual large language models by a marginal difference. Using fine-tuning methods like LoRA is effective for multilingual models; however, training the entire model on custom data remains effective. We found out that Flan T5 small model is the most effective when trained on our dataset. We infer the reason for this to be the small size of the model and the masked language modeling (MLM) pre-training method of this specific model.  

\section{Future Scope}
There's scarcity of amount of data for the any code switching task for any Indian language. We believe that future research to be done on such datasets. Fine-tuning the model has yielded impressive results, as evidenced by the improved BLEU scores. However, given the limitations of BLEU and inablility of capturing diverse textual representations of the same vocabulary, establishing a common ground for understanding becomes crucial. This step is essential for enhancing the creation of models that can comprehend various textual nuances and expressions, ensuring a more comprehensive and accurate evaluation of their performance. 








\vspace{15pt}

\end{document}